\documentclass[11pt]{article}

\usepackage[preprint]{acl}

\usepackage{subcaption}
\usepackage{booktabs}
\usepackage{multirow}
\usepackage{makecell}   % needed for \makecell
\usepackage{graphicx}
\usepackage{amssymb}   % needed for \mathbb

\usepackage{stfloats}

\hypersetup{
    colorlinks=true,
    linkcolor=[rgb]{0.1,0.3,0.6},
    citecolor=[rgb]{0.1,0.3,0.6},
    urlcolor=[rgb]{0.1,0.3,0.6}
}

\usepackage{times}
\usepackage{latexsym}
\usepackage{xspace}
\usepackage{graphicx}

\usepackage[T1]{fontenc}
\usepackage[utf8]{inputenc}

\usepackage{microtype}

\usepackage{inconsolata}

\usepackage{graphicx}

\usepackage{amsmath}

\title{%\modelname: %From Explicit to Latent Reasoning Guardrails
Robust and Efficient Guardrails with Latent Reasoning
}

\author{
  Siddharth Sai \quad Xiaofei Wen \quad Muhao Chen \\
  University of California, Davis \\
  \texttt{\{sai,xfwe,muhchen\}@ucdavis.edu}
}
\usepackage[textsize=tiny]{todonotes}

\newcommand{\modelname}{\textsc{CoLaGuard}\xspace}

\begin{document}
\maketitle
\begin{abstract}
Maintaining the safety of large language models (LLMs) is crucial as they are increasingly deployed in real-world applications. Existing safety guardrails typically rely on single-pass classification or, more recently, distilled reasoning. Reasoning-based guardrails significantly outperform classification-only baselines, but they incur substantial query latency and token overhead that make them impractical for high-throughput deployment. To address this challenge, we propose \modelname, a guardrail model that transfers multi-step safety reasoning into a continuous latent space through a stage-wise training curriculum, enabling direct hidden-state propagation at inference. Evaluated %across ten prompt and response moderation settings from eight unique safety benchmarks
on ten prompt- and response-moderation settings spanning eight safety benchmarks, \modelname improves macro-F1 by 8.24 points over Llama Guard 3 and matches our explicit reasoning baseline, GuardReasoner, in macro-F1 while delivering a 12.9$\times$ speedup and 22.4$\times$ reduction in token usage. Our results suggest that latent reasoning offers a practical alternative to explicit rationale generation for deployable guardrails, jointly improving safety robustness and inference efficiency rather than treating them as competing objectives. Project page: \href{https://github.com/SiddSai/Robust-and-Efficient-Guardrails-with-Latent-Reasoning}{github.com/SiddSai/CoLaGuard}

\end{abstract}
\section{Introduction}
\begin{figure*}[t]
    \centering
    \makebox[\textwidth][c]{%
        \includegraphics[width=1.00\textwidth]{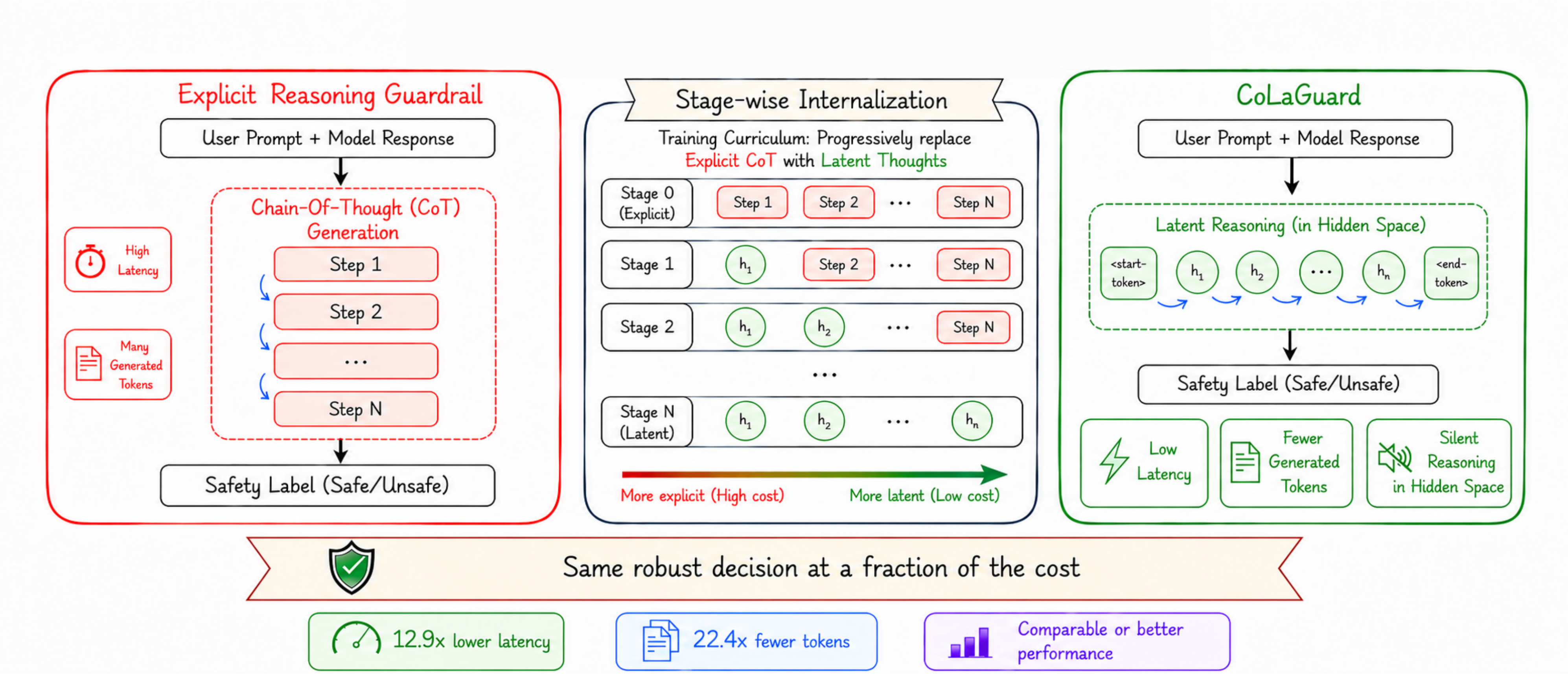}%
    }
    \caption{Overview of \modelname. Unlike explicit reasoning guardrails (left) that generate chain-of-thought tokens before assigning labels, \modelname (right) reasons through recurrent latent states, preserving moderation performance while avoiding token generation overhead and enabling 
12.9$\times$ faster inference and 22.4$\times$ fewer tokens. \modelname's stage-wise internalization curriculum (center) begins with explicit CoT supervision and progressively replaces reasoning tokens with latent states, shifting reasoning into hidden activations.}
    \label{fig:CoLaGuard-overview}
\end{figure*}

% Background & Motivation
As Large Language Models (LLMs) become integral to daily and industrial applications, ensuring their alignment with human values is critical. Although alignment training methods such as RLHF \citep{ouyang2022traininglanguagemodelsfollow, rafailov2023direct} can improve model behavior, they require modifying the target model and are costly to update after deployment. External safety guardrails \citep{inan2023llamaguardllmbasedinputoutput, han2024wildguard} therefore provide a practical alternative by offloading input and output moderation to smaller, third-party models. Early guardrails typically formulate moderation as single-pass classification, which is efficient but often becomes brittle under ambiguous, adversarial, or context-dependent safety decisions. 

Recent explicit reasoning guardrails~\citep{Wei2025thinkguard,liu2025guardreasonerreasoningbasedllmsafeguards} improve robustness by learning from distilled chain-of-thought (CoT) supervision~\citep{hsieh-etal-2023-distilling, kim2023cotcollectionimprovingzeroshot} and generating intermediate rationales before predicting a safety label. MrGuard~\citep{yang-etal-2025-mrguard} further extends reasoning-based guardrails to multilingual safety moderation by combining synthetic multilingual supervision with curriculum-guided Group Relative Policy Optimization (GRPO)~\citep{shao2024deepseekmathpushinglimitsmathematical}. However, this robustness comes at a steep computational cost. Because these models verbalize their intermediate rationales, moderation becomes a long autoregressive generation process. The additional CoT tokens substantially inflate inference time and completion-token cost, making explicit reasoning guardrails difficult to deploy in high-traffic, real-time settings~\citep{liu2025guardreasonerreasoningbasedllmsafeguards, sreedhar2025safetyreasoningempiricalstudy}. Existing efficiency-oriented variants, such as shorter supervised traces or reasoning on/off switches~\citep{Nvidia2025nemotron,sreedhar2025safetyreasoningempiricalstudy}, reduce the amount or frequency of rationale generation but still rely on explicit decoding and may sacrifice robustness.

% What I Did (Proposed Method)
This motivates a natural question: \textit{can guardrails retain the benefits of reasoning supervision without generating reasoning tokens at inference time?} We study this question through \modelname, a latent-reasoning safety guardrail that internalizes explicit safety rationales into continuous recurrent states as shown in Figure~\ref{fig:CoLaGuard-overview}.  Inspired by Coconut \citep{hao2025training} and ICoT-SI \citep{deng2025from}, \modelname %internalizes explicit safety rationales into a fixed number of latent recurrent steps. 
performs a fixed number of latent recurrent steps in place of explicit rationale generation.
It first learns from CoT supervision and then progressively replaces rationale tokens with latent states, allowing the model to directly predict the safety label without autoregressive rationale generation. A practical challenge is that pretrained LLMs are optimized to consume token embeddings rather than recirculated contextual hidden states, which can create a distribution mismatch during latent recurrence. To reduce this mismatch, we adopt Context-Prediction Fusion \citep{liu2026latentthoughtstuningbridging}, which combines contextual hidden-state information with predictive semantic guidance from the vocabulary embedding space. This stabilizes latent recurrence while preserving the latency and token-efficiency benefits of avoiding explicit CoT generation.

% Contributions List
% In summary, our contributions are:
% \begin{itemize}
%     \item We introduce \modelname, a latent-reasoning safety guardrail that internalizes explicit safety rationales through a stage-wise curriculum, enabling moderation without autoregressive rationale generation at inference time.

%    \item We demonstrate that \modelname preserves the robustness of explicit reasoning guardrails while substantially reducing inference cost, showing that reasoning-based moderation can be made practical for high-traffic deployment without autoregressive rationale generation.

%     \item We analyze the latent recurrence process and show that \modelname produces progressive safety-relevant representation shifts, while vanilla Coconut exhibits little change across recurrence steps.
% \end{itemize}

% Together, our results demonstrate that latent reasoning can preserve the robustness of explicit reasoning guardrails while delivering order-of-magnitude reductions in latency and token cost, marking a concrete step toward practical high-performance safety guardrails.

In summary, this work makes three main contributions. 
(1) We introduce \modelname, a latent-reasoning safety guardrail that internalizes explicit safety rationales through a stage-wise curriculum, enabling moderation without autoregressive rationale generation at inference time. 
(2) We show that \modelname preserves the robustness of explicit reasoning guardrails while substantially reducing inference cost, suggesting that reasoning-based moderation can be made practical without verbalized rationales. 
(3) We analyze the latent recurrence process and find that \modelname improves over vanilla Coconut, consistent with progressive safety-relevant representation shifts across latent steps that are largely absent in vanilla Coconut recurrence.

\section{Related Work}

\paragraph{LLM Guardrails} 
External guardrails provide a lightweight mechanism for safety moderation without modifying the base LLM. Early architectures such as Llama Guard~\citep{inan2023llamaguardllmbasedinputoutput} and WildGuard~\citep{han2024wildguard} treated moderation as classification, followed by models like ShieldGemma~\citep{zeng2024shieldgemmagenerativeaicontent}, Aegis~\citep{ghosh2024aegisonlineadaptiveai}, and Qwen3Guard~\citep{zhao2025qwen3guardtechnicalreport}, which improved performance through broader taxonomies. The broader guardrail literature has expanded robustness through adversarially resilient moderation~\citep{yuan2024rigorllm} and structured safety knowledge~\citep{kang2025rguard}. Recent work further improves performance with reasoning: GuardReasoner~\citep{liu2025guardreasonerreasoningbasedllmsafeguards} and ThinkGuard~\citep{Wei2025thinkguard} use chain-of-thought rationales~\citep{wei2023chainofthoughtpromptingelicitsreasoning} from expert models to improve generalization, while MrGuard~\citep{yang-etal-2025-mrguard} extends reasoning-based guardrails to multilingual moderation through synthetic multilingual supervision and curriculum-guided GRPO. Others explore efficiency trade-offs through shorter rationale traces and on/off switches~\citep{Nvidia2025nemotron,sreedhar2025safetyreasoningempiricalstudy,rebedea2023nemoguardrailstoolkitcontrollable}. However, because these models verbalize reasoning in natural language, they incur steep autoregressive decoding costs that limit their practicality for high-traffic, real-world deployment.

\paragraph{Latent Reasoning} 
A growing literature suggests that effective reasoning can occur within a model's hidden states rather than through explicit tokens \citep{chen2025reasoninglanguagecomprehensivesurvey, zhu2025surveylatentreasoning, biran2024hoppinglateexploringlimitations}. This space includes augmenting models with "thinking" tokens \citep{goyal2024think, zelikman2024quietstar, pfau2024letsthinkdotdot}, internalizing CoT through staged curricula \citep{deng2023implicitchainthoughtreasoning, deng2025from}, and feeding hidden states back as continuous input embeddings \citep{hao2025training, cheng2024compressedchainthoughtefficient, zhu2025surveylatentreasoning}. However, these methods have largely been studied on mathematical and logical reasoning tasks, and directly recycling raw hidden states can become unstable at larger scales due to distribution mismatch with the token embedding manifold. Latent Thoughts Tuning \citep{liu2026latentthoughtstuningbridging} addresses this with a context-prediction fusion mechanism that aligns contextual hidden states with predictive signals from the vocabulary embedding space. \modelname adapts these techniques, showing that latent reasoning can drastically reduce latency costs and preserve the robustness of explicit baselines in safety moderation.

% Methodology

\section{\modelname}
\label{sec:method}

We now present \modelname, a latent-reasoning guardrail framework for efficient prompt and response moderation. \modelname uses explicit safety rationales generated by expert models as training-time supervision, then progressively internalizes this step-by-step reasoning into recurrent latent states so that inference incurs only a fixed latent computation budget before decoding the safety labels. We formulate the guardrail task in \S\ref{sec:task}, describe reasoning-augmented supervision and explicit warm-up in \S\ref{sec:reasoning_supervision}--\S\ref{sec:warmup}, and present latent recurrence, stage-wise internalization and efficient inference in \S\ref{sec:latent_recurrence}--\S\ref{sec:inference}.

\subsection{Guardrail Task}
\label{sec:task}

Given a user prompt $x$ and a model response $s$, a guardrail model $G_{\theta}$ predicts the safety of both the input request and the generated response 
$(\hat{y}^{p}, \hat{y}^{r}) = G_{\theta}(x, s)$,
where $\hat{y}^{p} \in \mathcal{Y}$ denotes the predicted prompt harmfulness label, $\hat{y}^{r} \in \mathcal{Y}$ denotes the predicted response harmfulness label, and %$\mathcal{Y}=\{\texttt{harmful}, \texttt{unharmful}\}$.
$\mathcal{Y}$ denotes the set of safety categories in the guardrail's policy.

\subsection{Reasoning-Augmented Supervision}
\label{sec:reasoning_supervision}
The central challenge is maintaining the robustness of reasoning-based guardrails without requiring that the guardrail verbalize its reasoning process at inference time. \modelname addresses this by using explicit rationales for the initial training scaffolding. This follows prior work on chain-of-thought reasoning, step-by-step distillation, and reasoning-based safety guardrails, where intermediate rationales provide richer supervision than final labels alone~\citep{wei2023chainofthoughtpromptingelicitsreasoning,hsieh-etal-2023-distilling,kim2023cotcollectionimprovingzeroshot,liu2025guardreasonerreasoningbasedllmsafeguards,Wei2025thinkguard}.

We assume access to a reasoning-augmented guardrail corpus
$\mathcal{D} = \{(x_i, s_i, r_i, y_i)\}_{i=1}^{N}$,
where $x_i$ is a user prompt, $s_i$ is the corresponding model response, $y_i=(y_i^{p}, y_i^{r})$ contains the final prompt and response safety labels, and
\begin{equation*}
    r_i = (r_i^1, r_i^2, \ldots, r_i^{m_i})
\end{equation*}
is a step-separated safety rationale.

Unlike standard label-only guardrail training, this supervision exposes the model to the reasoning underlying the final moderation decision. However, \modelname does not aim to generate these rationales at inference time. Instead, the rationales serve as targets during the initial stages of training so that the model can later compress the deliberation process into latent steps.

\subsection{Explicit Reasoning Warm-Up}
\label{sec:warmup}

The first stage (Stage 0) trains the model as an explicit reasoning guardrail. Given an instruction $I$, prompt $x$, response $s$, rationale $r$, and final label tuple $y$, the model is optimized to generate structured safety-relevant rationale followed by the final safety labels:
\begin{equation*}
    \mathcal{L}_{\mathrm{warm}}
    =
    - \mathbb{E}_{(x,s,r,y)\sim\mathcal{D}}
    \log p_{\theta}(r, y \mid I, x, s).
    \label{eq:warmup_loss}
\end{equation*}
This warm-up follows the explicit reasoning guardrail paradigm, where models learn to verbalize intermediate safety reasoning before predicting final moderation labels \citep{liu2025guardreasonerreasoningbasedllmsafeguards, Wei2025thinkguard}. We denote the resulting model as $G_{\theta}^{0}$, from which subsequent stages progressively replace explicit rationale steps with latent recurrent steps.

%\subsection{Latent Recurrence with Context-Prediction Fusion}
\subsection{Dual-Mode Latent Recurrence}
\label{sec:latent_recurrence}

To internalize reasoning, \modelname switches between two modes. In language mode, the model consumes standard token embeddings and predicts the next token autoregressively. In latent mode, the model does not consume a standard token embedding; instead, the previous hidden state is fed back as the next input representation.

Let $e(\cdot)$ denote the token embedding function and let $h_t \in \mathbb{R}^{d}$ be the last-layer hidden state at position $t$. For a sequence with a latent span beginning at position $a$ and ending at position $b$, vanilla latent recurrence replaces the input embedding at each latent position with the previous hidden state:
\begin{equation*}
    E_t =
    \begin{cases}
        e(w_t), & t < a \text{ or } t > b, \\
        h_{t-1}, & a \leq t \leq b,
    \end{cases}
    \label{eq:latent_recurrence}
\end{equation*}
where $w_t$ is the discrete token at position $t$ outside the latent span. This follows the chain-of-continuous-thought formulation introduced by \citet{hao2025training}, allowing the model to perform recurrent computation in continuous latent space rather than generating intermediate rationale tokens.

While this latent recurrence lays the foundation of \modelname, directly feeding contextual hidden states back into a pretrained transformer creates a distribution mismatch since the base model is trained to consume token embeddings, while $h_{t-1}$ is a hidden representation. To reduce the hidden-state/token-embedding mismatch observed in latent recurrence, we adopt context-prediction fusion from Latent Thoughts Tuning \citep{liu2026latentthoughtstuningbridging}. At each latent position, the model first computes a predictive embedding from the next-token distribution induced by the previous hidden state:
\begin{equation*}
    e_{\mathrm{pred}}(h_{t-1})
    =
    \sum_{v \in \mathcal{V}_{p}}
    \tilde{p}_{\theta}(v \mid h_{t-1}) e(v),
    %\label{eq:pred_embedding}
\end{equation*}
where $\mathcal{V}_{p}$ is the nucleus-filtered vocabulary set, and $\tilde{p}_{\theta}(v \mid h_{t-1})$ is the renormalized probability distribution over this set. Structural latent-control tokens are excluded from this distribution.

The recurrent input is then constructed by fusing the contextual hidden state with the predictive embedding:
\begin{equation*}
    \tilde{e}_t
    =
    \alpha h_{t-1}
    +
    (1-\alpha)e_{\mathrm{pred}}(h_{t-1}),
    \label{eq:fusion_blend}
\end{equation*}
where $\alpha \in [0,1]$ controls the balance between contextual continuity and semantic anchoring. Finally, a lightweight projection module maps the fused representation back into the model input space:
\begin{equation*}
    \mathbf{e}^{\mathrm{in}}_t =
    \begin{cases}
        h_{t-1}, & \alpha = 1, \\
        g_{\phi}(\tilde{e}_t), & \alpha < 1 \text{ and adapter is used}, \\
        \tilde{e}_t, & \alpha < 1 \text{ and no adapter is used}.
    \end{cases}
    %\label{eq:fusion_adapter}
\end{equation*}
where $g_{\phi}$ is a trainable adapter. When $\alpha=1$, this reduces to vanilla hidden-state recurrence; when $\alpha<1$, the recurrent state is anchored by predictive information from the vocabulary embedding.

\subsection{Stage-Wise Internalization}
\label{sec:internalization}

The core of \modelname is a stage-wise curriculum that progressively replaces natural-language rationale steps with recurrent latent steps. The staged replacement schedule follows prior internalization curricula showing that gradually replacing explicit reasoning tokens is more stable than removing rationales all at once~\citep{deng2023implicitchainthoughtreasoning, deng2025from, hao2025training}.

For an example with $m$ rationale steps, we write the step-separated rationale as 

\begin{equation*}
    r = (r^1, r^2, \ldots, r^m),
\end{equation*}
let $K$ denote the maximum number of reasoning steps represented by the latent budget, and define $\ell_k=\min(k,K)$. At stage $k$, the first $k$ rationale steps are removed and replaced with $\ell_k c$ latent positions:
\begin{equation*}
    (r^1,\ldots,r^k)
    \rightarrow
    (z_1,\ldots,z_{\ell_k c}),
    \label{eq:stage_replacement}
\end{equation*}
where $c$ is the number of latent positions allocated per replaced reasoning step within the latent budget. We denote the resulting training sequence as $q^{(k)}$, which contains the instruction, prompt, response, latent span, any remaining rationale steps $r^{k+1}, \ldots, r^m$, and the final labels $y$.

If $k\geq m$, the rationale is fully replaced and the final label tuple follows the latent span directly; however, because the latent budget is fixed, examples with more than $K$ rationale steps may still contain explicit rationale tokens after the maximum latent stage is reached. We therefore include a final compression stage that keeps the latent span fixed at $Kc$ positions while removing all remaining rationale steps:
\begin{equation*}
    r^{1:m} \rightarrow (z_1,\ldots,z_{Kc}).
    \label{eq:compression_replacement}
\end{equation*}
This extra stage enables the absorption of residual explicit reasoning signal into the fixed latent recurrence rather than remaining decoded as text.

Training optimizes only the remaining language tokens and final labels. The prompt, response, latent-control tokens, and latent positions are masked from the language-modeling loss. Let $\mathcal{M}^{(k)}$ be the set of supervised token positions in $q^{(k)}$. The internalization objective is
\begin{equation*}
    \mathcal{L}_{\mathrm{int}}^{(k)}
    =
    -
    \mathbb{E}_{(x,s,r,y)\sim\mathcal{D}}
    \sum_{t \in \mathcal{M}^{(k)}}
    \log p_{\theta}(q^{(k)}_t \mid q^{(k)}_{<t}).
    \label{eq:internalization_loss}
\end{equation*}
The same masked language-modeling objective is used for the final compression stage, with supervised positions restricted to the final safety-label tokens.

As $k$ increases, less of the original rationale remains in language space, forcing more of the safety decision process to be represented by latent recurrence. This objective gives the latent positions no direct textual target so that the latent states are optimized through their downstream ability to predict the remaining rationale steps and the final safety labels.

\subsection{Efficient Inference}
\label{sec:inference}

At inference time, \modelname receives only the instruction, prompt, and response. It appends a fixed latent span and performs recurrent latent computation using the fused update in \S\ref{sec:latent_recurrence}. Let
%\begin{equation*}
$
    Z_L =
    \big[
    \langle \mathrm{start\text{-}latent} \rangle,
    z_1,\ldots,z_L,
    \langle \mathrm{end\text{-}latent} \rangle
    \big]
$
%\end{equation*}
denote the fixed latent span, where $L$ is the latent budget used at deployment. After the latent span, the model returns to language mode and autoregressively predicts the final prompt and response safety labels:
\begin{equation*}
    (\hat{y}^{p}, \hat{y}^{r})
    =
    \arg\max_{(y^{p},y^{r}) \in Y^2}
    p_{\theta}(y^{p}, y^{r} \mid I, x, s, Z_L).
    \label{eq:inference}
\end{equation*}

Because \modelname does not generate natural-language rationales, its inference cost scales with the number of latent positions rather than the length of an explicit chain-of-thought.

\begin{table}[t]
\centering
\footnotesize
\begin{tabular}{lr}
\toprule
\textbf{Benchmark} & \textbf{Samples} \\
\midrule
\multicolumn{2}{c}{\textit{Prompt Harmfulness Detection}} \\
\midrule
ToxicChat~\citep{lin2023toxicchatunveilinghiddenchallenges}           & 2,853 \\
OpenAI Moderation~\citep{markov2023holisticapproachundesiredcontent} & 1,680 \\
Aegis Safety Test~\citep{ghosh2024aegisonlineadaptiveai}     &   359 \\
HarmBench~\citep{mazeika2024harmbenchstandardizedevaluationframework}       &   239 \\
WildGuardTest~\citep{han2024wildguard}       & 1,756 \\
\midrule
\multicolumn{2}{c}{\textit{Response Harmfulness Detection}} \\
\midrule
HarmBench~\citep{mazeika2024harmbenchstandardizedevaluationframework}       &   602 \\
SafeRLHF~\citep{ji2024pkusaferlhf}           & 2,000 \\
BeaverTails~\citep{ji2023beavertails}        & 3,021 \\
XSTest~\citep{rottger-etal-2024-xstest}      &   446 \\
WildGuardTest~\citep{han2024wildguard}       & 1,768 \\
\bottomrule
\end{tabular}
\caption{Evaluation benchmarks for prompt and response harmfulness detection.}
\label{tab:eval_benchmarks}
\end{table}

\begin{table*}[t]
\centering
\caption{F1 Score (\%) of Models on 5 Benchmarks of Prompt Harmfulness Detection.
  \textbf{Bold} and \underline{underlined} values denote the best and runner-up.
  ``--'' denotes the result is unavailable.}
\label{tab:prompt_harmfulness}
%\resizebox{\textwidth}{!}{%
\setlength{\tabcolsep}{4pt}
\footnotesize
\begin{tabular}{llccccccc}
\toprule
\textbf{Method} & \textbf{\makecell{Model\\Size}} &
\textbf{ToxicChat} & \textbf{HarmBench} &
\textbf{\makecell{OpenAI\\Mod.}} &
\textbf{\makecell{Aegis\\SafetyTest}} &
\textbf{\makecell{WildGuard\\Test}} &
\textbf{\makecell{Macro\\Avg}} &
\textbf{\makecell{Micro\\Avg}} \\
\midrule
\multicolumn{9}{c}{\textit{Closed-Source Guard API}} \\
\midrule
GPT-4o             & Unknown & 64.46 & 82.27 & 62.26 & 81.07 & 80.87 & 74.19 & 69.59 \\
GPT-4o+CoT         & Unknown & 73.43 & 81.98 & 76.78 & 88.24 & 82.75 & 80.64 & 77.69 \\
o1-preview         & Unknown & 57.69 & 89.61 & 74.60 & 83.15 & 76.31 & 76.27 & 69.00 \\
\midrule
\multicolumn{9}{c}{\textit{Open-Source Guard Model}} \\
\midrule
LLaMA Guard            & 7B  & 61.60 & 67.20 & 75.80 & 74.10 & 56.00 & 66.94 & 64.48 \\
LLaMA Guard 2          & 8B  & 47.10 & 94.00 & 76.10 & 71.80 & 70.90 & 71.98 & 63.16 \\
LLaMA Guard 3          & 8B  & 53.12 & \textbf{98.94} & 79.69 & \textbf{99.50} & 68.47 & 79.94 & 67.52 \\
Aegis Guard Defensive  & 7B  & 70.00 & 77.70 & 67.50 & 84.80 & 78.50 & 75.70 & 72.60 \\
Aegis Guard Permissive & 7B  & 73.00 & 70.50 & 74.70 & 82.90 & 71.50 & 74.52 & 73.46 \\
Aegis Guard 2.0        & 8B  & --    & --    & 81.00 & --    & 81.60 & --    & --    \\
ShieldGemma            & 2B  &  6.91 & 11.81 & 13.89 &  7.47 &  9.36 &  9.89 &  9.44 \\
ShieldGemma            & 9B  & 67.92 & 67.96 & 78.58 & 77.63 & 57.74 & 69.97 & 68.43 \\
WildGuard              & 7B  & 70.80 & \underline{98.90} & 72.10 & 89.40 & 88.90 & 84.02 & 77.68 \\
QwQ-preview            & 32B & 34.81 & 86.73 & 61.58 & 80.23 & 66.02 & 65.87 & 53.47 \\
GuardReasoner          & 1B  & 72.43 & 96.31 & 70.06 & 89.34 & 87.37 & 83.10 & 77.37 \\
GuardReasoner          & 3B  & \underline{78.20} & 89.10 & 71.87 & \underline{91.39} & 89.01 & 83.91 & \underline{80.48} \\
GuardReasoner          & 8B  & \textbf{78.79} & 91.86 & 72.00 & 90.18 & \underline{89.17} & \textbf{84.40} & \textbf{80.83} \\
\midrule
\multicolumn{9}{c}{\textit{Latent Reasoning Guardrail (Ours)}} \\
\midrule
\modelname (Ours)       & 3B  & 75.27 & 94.25 & 73.15 & 90.58 & 88.15 & \underline{84.28} & 79.49 \\
\modelname (Ours)       & 8B  & 75.26 & 93.54  & 73.45 & 89.45 & \textbf{89.44} & 84.23 & 79.77 \\
\bottomrule
\end{tabular}%
%}
\end{table*}

\section{Experiments}

To evaluate \modelname, we conduct experiments on multiple safety benchmarks, comparing (1) safety classification performance across various baselines and (2) inference efficiency against explicit reasoning guardrails. 

\subsection{Experimental Setup}

\paragraph{Reasoning Augmented Dataset.} We use the GuardReasonerTrain dataset \citep{liu2025guardreasonerreasoningbasedllmsafeguards} as the primary training source for our guardrail model. GuardReasonerTrain is a 127,000-example reasoning-augmented compilation of the following safety-focused datasets: WildGuard~\citep{han2024wildguard}, AegisSafety~\citep{ghosh2024aegisonlineadaptiveai}, BeaverTails~\citep{ji2023beavertails}, and ToxicChat~\citep{lin2023toxicchatunveilinghiddenchallenges}. Each example comes with a prompt, composed of the guardrail instructions, a user input, and a user output; a multi-step reasoning trace separated into the three tasks of request moderation, refusal detection, and response moderation; and ground-truth answers for the three tasks. Using the same reasoning-augmented supervision source as GuardReasoner  allows us to directly compare explicit rationale generation against latent internalization under a matched training signal.

Both iCoT~\citep{deng2025from} and Coconut~\citep{hao2025training} show that replacing too many language tokens per stage can destabilize training. We therefore split reasoning traces into smaller step-level replacements, but this increases the number of training stages and overall computational cost. To reduce cost and focus on request and response safety moderation, we remove the refusal task from \modelname training supervision. 

\paragraph{Training Details.}
We use separate training configurations for the explicit CoT warm-up stage and the latent internalization stages. In Stage 0, we fully fine-tune Llama 3.1 8B on GuardReasonerTrain to obtain an explicit reasoning baseline. Training is performed on 8$\times$A100 (80GB) GPUs for 3 epochs, with per-device batch size 1, gradient accumulation 32, AdamW optimization~\citep{loshchilov2019decoupledweightdecayregularization}, a cosine learning-rate schedule, and an initial learning rate of $5\times10^{-5}$. The fusion coefficient is set to $\alpha=1.0$ in this stage, so the fusion module is inactive.

\begin{table*}[t]
\centering
\caption{F1 Score (\%) of Models on 5 Benchmarks of Response Harmfulness Detection.
  \textbf{Bold} and \underline{underlined} values denote the best and runner-up.
  ``--'' denotes the result is unavailable.}
\label{tab:response_harmfulness}
%\resizebox{\textwidth}{!}{%
\footnotesize
\setlength{\tabcolsep}{4pt}
\begin{tabular}{llccccccc}
\toprule
\textbf{Method} & \textbf{\makecell{Model\\Size}} &
\textbf{HarmBench} & \textbf{SafeRLHF} & \textbf{BeaverTails} &
\textbf{\makecell{XSTest\\Response}} &
\textbf{\makecell{WildGuard\\Test}} &
\textbf{\makecell{Macro\\Avg}} &
\textbf{\makecell{Micro\\Avg}} \\
\midrule
\multicolumn{9}{c}{\textit{Closed-Source Guardrail API}} \\
\midrule

GPT-4o             & Unknown & 56.34 & 64.05 & 78.63 & 65.12 & 65.24 & 65.88 & 69.41 \\
GPT-4o+CoT         & Unknown & 65.99 & 65.10 & 82.26 & 86.90 & 71.43 & 74.34 & 74.45 \\
o1-preview         & Unknown & 76.40 & 66.60 & 79.96 & 74.75 & 50.00 & 69.54 & 69.22 \\

\midrule
\multicolumn{9}{c}{\textit{Open-Source Guardrail}} \\
\midrule
LLaMA Guard            & 7B  & 52.00 & 48.40 & 67.10 & 82.00 & 50.50 & 60.00 & 58.27 \\
LLaMA Guard 2          & 8B  & 77.80 & 51.60 & 71.80 & 90.80 & 66.50 & 71.70 & 66.99 \\
LLaMA Guard 3          & 8B  & 85.07 & 44.36 & 67.84 & 87.67 & 70.80 & 71.15 & 64.97 \\
Aegis Guard Defensive  & 7B  & 62.20 & 59.30 & 74.70 & 52.80 & 49.10 & 59.62 & 62.79 \\
Aegis Guard Permissive & 7B  & 60.80 & 55.90 & 73.80 & 60.40 & 56.40 & 61.46 & 63.55 \\
Aegis Guard 2.0        & 8B  & --    & --    & --    & 86.20 & 77.50 & --    & --    \\
ShieldGemma            & 2B  & 35.36 & 16.92 & 30.97 & 65.55 & 20.13 & 33.79 & 27.24 \\
ShieldGemma            & 9B  & 56.44 & 47.07 & 63.61 & 73.86 & 47.00 & 57.60 & 55.67 \\
HarmBench LLaMA        & 13B & 84.30 & 60.00 & 77.10 & 64.50 & 45.70 & 66.32 & 65.49 \\
HarmBench Mistral      & 7B  & \textbf{87.00} & 52.40 & 75.20 & 72.00 & 60.10 & 69.34 & 66.70 \\
MD-Judge               & 7B  & 81.60 & 64.70 & 86.70 & 90.40 & 76.80 & 80.04 & 78.67 \\
BeaverDam              & 7B  & 58.40 & \textbf{72.10} & \textbf{89.90} & 83.60 & 63.40 & 73.48 & 76.60 \\
WildGuard              & 7B  & 86.30 & 64.20 & 84.40 & \textbf{94.70} & 75.40 & 81.00 & 77.95 \\
QwQ-preview            & 32B & 69.65 & 62.76 & 77.26 & 45.95 & 17.56 & 54.64 & 57.73 \\
GuardReasoner          & 1B  & 84.75 & 68.39 & 85.84 & 90.12 & 74.81 & 80.78 & 79.06 \\
GuardReasoner          & 3B  & 85.66 & 69.02 & 86.72 & 91.36 & \underline{79.70} & 82.49 & 80.80 \\
GuardReasoner          & 8B  & 85.47 & 70.04 & \underline{87.60} & \underline{94.34} & 78.20 & \underline{83.13} & \underline{81.22} \\
\midrule
\multicolumn{9}{c}{\textit{Latent Reasoning Guardrail (Ours)}} \\
\midrule
\modelname       & 3B  & 86.36 & 68.72 & 86.29 & 94.19 & 77.23 & 82.56 & 80.22 \\
\modelname       & 8B  & \underline{86.38} & \underline{70.49} & 86.55 & 92.02 & \textbf{81.23} & \textbf{83.33} & \textbf{81.55} \\
\bottomrule
\end{tabular}%
%}
\end{table*}

Starting from the Stage-0 checkpoint, we then train the stage-wise internalization curriculum. Since roughly 80\% of GuardReasonerTrain examples contain at most six reasoning steps, we use six latent recurrent steps as the fixed inference budget. Each internalization stage replaces one additional reasoning step with latent states, and a final compression stage removes any remaining explicit reasoning for longer traces while preserving the same six-step latent budget. Each stage is trained for one epoch with a reset AdamW optimizer and a constant learning rate of $1\times10^{-5}$.

During internalization, we linearly anneal the fusion coefficient from $\alpha=1.0$ to $\alpha=0.6$ over the first 200 warm-up steps. We set the fusion temperature to 1.0, top-$p$ to 0.9, and use a fusion adapter with hidden dimension 1024. All training is conducted in bf16 precision, and checkpoints are saved after each stage.

For the 3B model, we use Llama 3.2 3B as the backbone and set the internalization learning rate to $2\times10^{-5}$, which we found more stable for this scale. Following the implementation choice in \citet{liu2026latentthoughtstuningbridging}, we disable the fusion adapter for the 3B model because Llama 3.2 3B uses tied input-output embeddings. All other training settings are kept identical to the 8B configuration.

\paragraph{Safety Evaluation.} To assess the performance and efficiency of our guardrail model while isolating the effect of latent reasoning against explicit rationale generation under matched supervision, we evaluate on benchmarks used by \citet{liu2025guardreasonerreasoningbasedllmsafeguards} (Table~\ref{tab:eval_benchmarks}) and use GuardReasoner (SFT-only, without hard-sample DPO) as our primary explicit reasoning baseline. More details on these benchmarks can be found in Appendix~\ref{sec:safeval}.

We compare \modelname against 20 baselines spanning closed-source APIs, open-source guard models, and our primary explicit reasoning baseline. Baseline names and model sizes are reported in Tables~\ref{tab:prompt_harmfulness} and~\ref{tab:response_harmfulness}; corresponding references are provided in Appendix~\ref{sec:baseline_details}.

\subsection{Results}

% Prompt Harmfulness Detection Table — CoLaGuard 3B filled in (checkpoint 11)
% Bold = best per column, underline = runner-up per column
% Micro Avg weighted by: TC=2853, HB=239, OAI=1680, Aegis=359, WG=1756 (total=6887)

% Response Harmfulness Detection Table — CoLaGuard 3B filled in (checkpoint 11)
% Bold = best per column, underline = runner-up per column
% Micro Avg weighted by: HB=602, SRLHF=2000, BT=3021, XST=446, WG=1768 (total=7837)

\paragraph{Overall Classification Performance.}
Tables~\ref{tab:prompt_harmfulness} and~\ref{tab:response_harmfulness} report F1 scores on prompt and response harmfulness detection. \modelname~8B is comparable to GuardReasoner~8B, with prompt macro-F1 of 84.23 vs.\ 84.40 and response macro-F1 of 83.33 vs.\ 83.13. Compared with Llama Guard~3, it improves the average macro-F1 across both tasks by 8.24 points while avoiding explicit rationale generation.

At the benchmark level, \modelname~8B achieves the best F1 on WildGuardTest for both prompt and response detection (89.44 and 81.23), and ranks second on HarmBench response and SafeRLHF (86.38 and 70.49). Its lower prompt micro-F1 relative to GuardReasoner~8B (79.77 vs.\ 80.83) is mainly due to ToxicChat, which accounts for 41.4\% of the prompt evaluation set and therefore has a large effect on the micro average.

\paragraph{Model Size Comparison.}
\modelname~3B is already competitive with GuardReasoner~3B, slightly improving both prompt macro-F1 (84.28 vs.\ 83.91) and response macro-F1 (82.56 vs.\ 82.49). Scaling to 8B mainly benefits response detection and yields better combined averages (83.78 vs.\ 83.42 macro; 80.66 vs.\ 79.86 micro), suggesting a modest but more consistent gain from the larger backbone.

\begin{table*}[t]
\centering
\footnotesize
\setlength{\tabcolsep}{4pt}
\caption{Inference Efficiency and Performance Comparison. We report inference time, completion token cost, and combined macro-F1 (averaged across prompt and response moderation). Inference is conducted on 1$\times$H100 (80GB) GPU. LlamaGuard-3-8B is a single-pass, non-reasoning classifier included as a lower bound on latency and token cost.}
\label{tab:efficiency}
\begin{tabular}{lccccc}
\toprule
\multirow{2}{*}{\textbf{Metric}} 
  & \multicolumn{2}{c}{\textbf{3B}} 
  & \multicolumn{3}{c}{\textbf{8B}} \\
\cmidrule(lr){2-3} \cmidrule(lr){4-6}
& GuardReasoner & \textsc{CoLaGuard} 
& GuardReasoner & \textsc{CoLaGuard} & LlamaGuard-3 \\
\midrule
Time Cost (ms/query)      & 3801.03 & 318.9  & 4407.8 & 342.0  & 110.56 \\
Token Cost (token/query)  & 281.96  & 13.0   & 289.4  & 12.9   & 3.67   \\
Macro-F1 (Combined)       & 83.20   & 83.42  & 83.77  & 83.78  & 75.55  \\
\bottomrule
\end{tabular}
\end{table*}

\paragraph{Inference Efficiency.}
Table~\ref{tab:efficiency} shows that \modelname substantially reduces inference cost compared with GuardReasoner. At 8B, latency drops from 4{,}407.8 to 342.0~ms/query, a 12.9$\times$ speedup, while token usage decreases from 289.4 to 12.9~tokens/query, a 22.4$\times$ reduction, with combined macro-F1 held essentially constant (83.77 vs.\ 83.78). These gains come from replacing long autoregressive CoT generation with a fixed six-step latent recurrence. We additionally report LlamaGuard-3-8B, a single-pass non-reasoning classifier, as a lower bound on latency and token cost. It is faster than \modelname (110.56~ms/query, 3.67~tokens/query) but trails its combined macro-F1 by 8.24 points (75.55 vs.\ 83.78), confirming that \modelname's efficiency gains reflect a genuine accuracy trade-off against reasoning-based guardrails rather than a shortcoming relative to non-reasoning classifiers.

\subsection{Ablation Studies}
\paragraph{Analyzing Latent Recurrence Dynamics.}
\begin{figure}[t]
    \centering
    \includegraphics[width=\columnwidth]{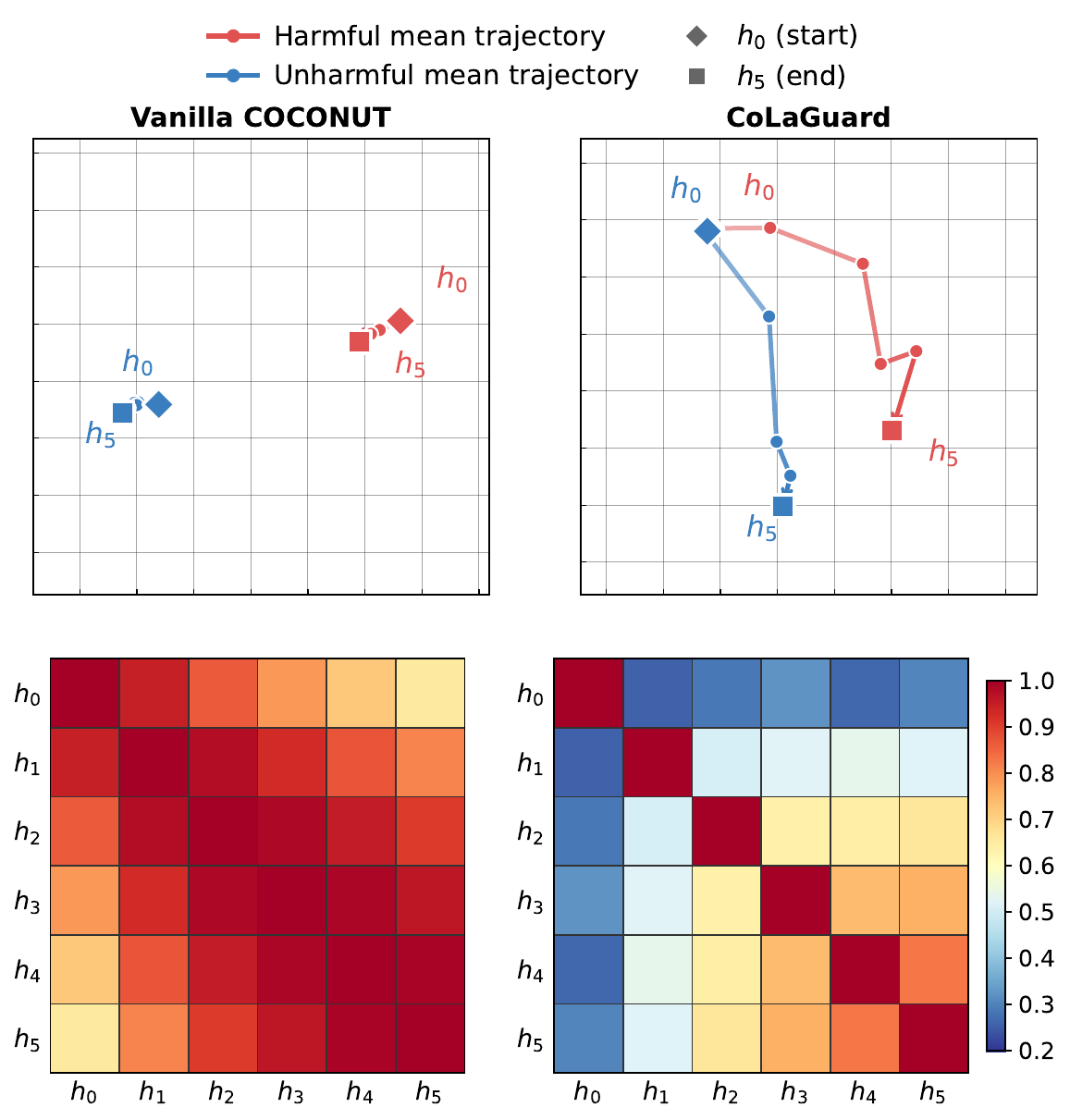}
    \caption{Geometric Analysis of Latent Representations. \textit{(Top)} UMAP of mean harmful/unharmful trajectories across recurrence steps $h_0$--$h_5$. \textit{(Bottom)} Intra-sample cosine similarity heatmap between latent steps. Vanilla Coconut shows highly similar latent states and early label separation, while \textsc{CoLaGuard} exhibits progressive class differentiation across recurrence steps.}
    \vspace{-0.5em}
    \label{fig:CoLaGuard-analysis}
\end{figure}
Recent work questions whether latent tokens in Coconut-style recurrence perform meaningful computation beyond acting as learned placeholders. \citet{zhang2025latenttokensthinkcausal} find that vanilla Coconut tokens form clustered embeddings with limited input sensitivity, suggesting placeholder behavior from learned shortcuts. \citet{liu2026latentthoughtstuningbridging} show that Context-Prediction Fusion mitigates inter-sample representational collapse, suggesting more expressive latent states. We extend this analysis to \modelname through WildGuardTest latent trajectories and a full-suite Context-Prediction-Fusion ablation against vanilla Coconut.

Figure~\ref{fig:CoLaGuard-analysis} shows the average pairwise cosine similarity between latent steps $(h_i, h_j)$ across samples and mean harmful/unharmful trajectories via UMAP ~\citep{mcinnes2020umapuniformmanifoldapproximation}. Vanilla Coconut exhibits uniformly high cross-step similarity, consistent with early commitment to a fixed latent state that is simply propagated forward; its harmful and unharmful trajectories are already separated at $h_0$, with limited additional separation in later steps. In contrast, \modelname shows noticeably lower cross-step similarity, indicating that its latent states continue to evolve throughout the recurrence rather than collapsing after the initial step. Its trajectories begin closer together and diverge progressively, suggesting that recurrence contributes to the refinement of safety-relevant representations rather than simply preserving an early decision.

As an ablation of Context-Prediction Fusion, a vanilla Coconut guardrail with the same six-step latent budget reaches 81.82 combined macro-F1 and 79.78 combined micro-F1, compared with 83.78 and 80.72 for \modelname. Context-Prediction Fusion yields clear gains that bring it to parity with the explicit reasoning baseline (+1.96 macro-F1, +0.94 micro-F1), suggesting that the more progressive latent shifts in Figure~\ref{fig:CoLaGuard-analysis} may be relevant to downstream moderation performance.

Furthermore, the complementary interventional check in Appendix~\ref{sec:budget_sweep} reports a latent-budget sweep on the same checkpoint, showing that \modelname's combined macro-F1 increases monotonically with the number of latent steps used at inference, from 80.92 at zero steps to 83.78 at the full six-step budget.

% \paragraph{Tradeoffs Between Accuracy and Inference Efficiency.}

\paragraph{Scaling Training Data.}
\begin{figure}[t]
    \centering
    \includegraphics[width=\columnwidth,  trim=0 20 0 0, clip]{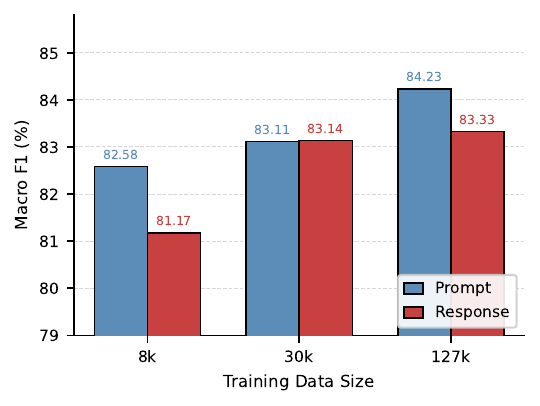}
    \caption{Training Data Scaling. \modelname~8B prompt and 
    response macro-F1 across training data sizes.}
    \label{fig:scaling}
    \vspace{-0.5em}
\end{figure}

Figure~\ref{fig:scaling} shows that \modelname~8B improves consistently with more reasoning-augmented training data. Response macro-F1 improves sharply from 8k to 30k examples (+1.97 points) but shows limited additional gain at 127k (+0.19 points). Prompt macro-F1 increases more gradually, gaining 0.53 points from 8k to 30k and 1.12 points from 30k to 127k.

These trends suggest that response moderation benefits earlier from diverse supervision, while prompt moderation continues to improve at larger scale. Overall, the results show that \modelname scales reliably with training data and achieves its best performance with the full GuardReasonerTrain corpus.

\section{Conclusion}
We introduced \modelname, a latent reasoning guardrail that internalizes explicit safety reasoning through a stage-wise curriculum. Across prompt and response harmfulness detection benchmarks, \modelname matches the average macro-F1 of an explicit reasoning guardrail while substantially reducing inference cost. \modelname~8B matches GuardReasoner~8B in macro-F1 while achieving 12.9$\times$ lower latency and 22.4$\times$ fewer tokens. These results show that latent reasoning is a practical path toward safety guardrails that are both robust and efficient for deployment.

% Work in Progress
\section*{Limitations}
While \modelname demonstrates strong efficiency and competitive safety performance, several limitations remain. First, our evaluation focuses on text-based prompt and response harmfulness detection, leaving broader policy taxonomies, multilingual inputs, multimodal content, adaptive adversaries, and long-horizon agent behavior for future work. Second, \modelname is trained from distilled reasoning traces and may inherit biases or coverage gaps from the underlying supervision, which latent internalization may make harder to detect than in models that still verbalize their reasoning. Finally, although our latent representation analysis suggests progressive safety-relevant refinement, more causal interventions are needed to fully characterize how each latent step contributes to the final decision to improve interpretability of safety decisions. 

% Work in Progress
\section*{Ethics Statement}
The aim of this work is to improve the reliability and efficiency of LLM safety guardrails. While latent reasoning moderation may make strong safety filters more practical in high-traffic settings, these guardrails can still produce false positives and false negatives on ambiguous or context-dependent inputs. Therefore, \modelname itself should not be considered a replacement for human oversight in real-world deployment, but rather should be used as part of a broader moderation system. The safety data used in the evaluation and training processes may contain harmful or sensitive content and should be handled with appropriate access controls and annotator-care practices.


% REFERENCES SECTION
\bibliography{custom}
\clearpage
\appendix

% \section{Appendix}
% \label{sec:appendix}

\section{Safety Evaluation}
\label{sec:safeval}

\subsection{Description of Benchmarks}
To assess the performance and efficiency of our latent reasoning guardrail model, we evaluate it across eight unique safety-related benchmarks.

\emph{WildGuard}~\citep{han2024wildguard}: WildGuardMix is a large-scale safety moderation dataset with 92,000 labeled examples that cover both normal and adversarial prompt behaviors that come coupled with corresponding refusal and compliance responses. The WildGuardTest split is human-annotated and covers 5,000 safety labeled examples.

\emph{ToxicChat}~\citep{lin2023toxicchatunveilinghiddenchallenges}: ToxicChat is a benchmark that includes 10,000 real user queries, leveraged as adversarial prompts for testing content moderation and toxicity detection in human-AI interactions.

\emph{Aegis Safety Test 1.0}~\citep{ghosh2024aegisonlineadaptiveai}: A dataset of approximately 11,000 manually annotated examples, Aegis Safety Test 1.0 was curated with the purpose of testing LLM safety alignment in accordance with Nvidia's content safety taxonomy.

\emph{HarmBench}~\citep{mazeika2024harmbenchstandardizedevaluationframework}: HarmBench is a framework that is systematically designed to address the lack of standardized evaluation frameworks in the field of automated red teaming. By leveraging various behaviors, this framework can be used to generate red-teaming test cases for evaluating the adversarial robustness of LLMs.

\emph{OpenAI Moderation}~\citep{markov2023holisticapproachundesiredcontent}: A benchmark for assessing LLMs' ability to detect harmful content based on OpenAI's safety guidelines, covering violence, self-harm, and misinformation. 

\emph{SafeRLHF}~\citep{ji2024pkusaferlhf}: A dataset of 82,000 questions with two responses each, every entry in SafeRLHF includes safety meta-labels as well as preference between the two responses.

\emph{BeaverTails}~\citep{ji2023beavertails}: The Beavertails dataset was introduced to further research on safety alignment in LLMs. The complete dataset includes over 300,000 question-answer pairs that are annotated with safety meta-labels and corresponding, violated safety categories. 

\emph{XSTest}~\citep{rottger-etal-2024-xstest}: Developed to evaluate refusal behaviors and identify systematic failure modes in large language models, XSTest is comprised of 250 safe prompts across ten prompt types and contrasting 200 unsafe prompts that human-aligned models should refuse.

\subsection{Baseline Details}
\label{sec:baseline_details}

Details are presented in Table~\ref{tab:baseline_details}.

\begin{table}[t]
\centering
\small
\begin{tabular}{lll}
\toprule
Baseline & Reference & Model Size \\
\midrule
GPT-4o & \citet{openai2024o1} & Unknown \\
GPT-4o + CoT & \citet{openai2024o1} & Unknown \\
o1-preview & \citet{openai2024o1} & Unknown \\
LLaMA Guard & \citet{inan2023llamaguardllmbasedinputoutput} & 7B \\
LLaMA Guard 2 & \citet{metallamaguard2} & 8B \\
LLaMA Guard 3 & \citet{grattafiori2024llama3} & 8B \\
Aegis Guard Defensive & \citet{ghosh2024aegisonlineadaptiveai} & 7B \\
Aegis Guard Permissive & \citet{ghosh2024aegisonlineadaptiveai} & 7B \\
Aegis Guard 2.0 & \citet{ghosh2025aegis2} & 8B \\
ShieldGemma & \citet{zeng2024shieldgemmagenerativeaicontent} & 2B / 9B \\
WildGuard & \citet{han2024wildguard} & 7B \\
QwQ-preview & \citet{qwq2024preview} & 32B \\
HarmBench LLaMA & \citet{mazeika2024harmbenchstandardizedevaluationframework} & 13B \\
HarmBench Mistral & \citet{mazeika2024harmbenchstandardizedevaluationframework} & 7B \\
MD-Judge & \citet{li2024saladbenchhierarchicalcomprehensivesafety} & 7B \\
BeaverDam & \citet{ji2023beavertails} & 7B \\
GuardReasoner & \citet{liu2025guardreasonerreasoningbasedllmsafeguards} & 1B / 3B / 8B \\
\bottomrule
\end{tabular}
\caption{Baseline references and model sizes for Tables~\ref{tab:prompt_harmfulness} and~\ref{tab:response_harmfulness}.}
\label{tab:baseline_details}
\end{table}

\subsection{Latent Budget Sweep}
\label{sec:budget_sweep}

\begin{figure}[h]
    \centering
    \includegraphics[width=0.9\columnwidth]{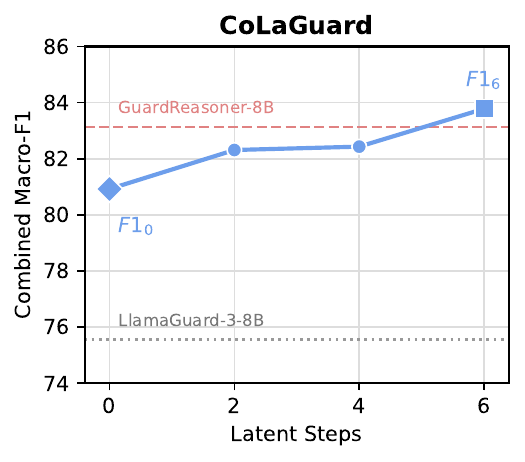}
    \caption{Latent budget sweep. Combined macro-F1 of \modelname~8B increases monotonically with the number of latent recurrence steps used at inference, from 80.92 at zero steps to 83.78 at the full six-step budget, closing the gap to GuardReasoner-8B (dashed) while remaining well above the non-reasoning LlamaGuard-3-8B floor (dotted).}
    \label{fig:budget_sweep}
\end{figure}

To test whether latent recurrence performs computation that is functionally load-bearing rather than acting as a fixed placeholder, we evaluate the same \modelname~8B checkpoint at inference-time latent budgets of 0, 2, 4, and 6 steps. We take the six-step checkpoint and truncate the latent span at inference, allowing the model to decode safety labels after the truncated number of recurrent steps. Figure~\ref{fig:budget_sweep} reports combined macro-F1 (averaged across the prompt and response moderation benchmarks in Tables~\ref{tab:prompt_harmfulness} and~\ref{tab:response_harmfulness}) at each budget, with GuardReasoner-8B and LlamaGuard-3-8B included as reference points.

Combined macro-F1 increases monotonically with step count, from 80.92 at zero latent steps to 83.78 at the full six-step budget. Notably, \modelname already outperforms LlamaGuard-3-8B by 5.37 points at zero latent steps, indicating that reasoning-augmented training alone provides a strong accuracy baseline independent of latent recurrence. Each additional latent step then contributes a further, roughly monotonic gain, closing the remaining gap to GuardReasoner-8B. This provides interventional evidence, complementary to the representational analysis in Figure~\ref{fig:CoLaGuard-analysis}, that the latent recurrence is not redundant and that each step contributes meaningfully to the final safety decision.

Table~\ref{tab:budget_sweep} additionally reports the corresponding latency at each budget. Latency increases only modestly across the sweep, from 236.34~ms/query at zero latent steps to 342.00~ms/query at six steps, remaining an order of magnitude below GuardReasoner-8B's 4,407.8~ms/query throughout, so the accuracy gains from additional latent steps come at comparatively little additional inference cost.

\begin{table}[h]
\centering
\small
\begin{tabular}{lcc}
\toprule
\textbf{Latent Steps} & \textbf{Time (ms/query)} & \textbf{Macro-F1} \\
\midrule
LlamaGuard-3-8B & 110.56 & 75.55 \\
0  & 236.34 & 80.92 \\
2  & 255.77 & 82.31 \\
4  & 293.63 & 82.43 \\
6  & 342.00 & 83.78 \\
GuardReasoner-8B & 4407.8 & 83.13 \\
\bottomrule
\end{tabular}
\caption{Latency and combined macro-F1 across latent budgets for \modelname~8B, with LlamaGuard-3-8B and GuardReasoner-8B as reference points.}
\label{tab:budget_sweep}
\end{table}

\end{document}